\documentclass{article}
\usepackage[utf8]{inputenc}
\usepackage[preprint]{IIBPSL_2020}
\usepackage{graphicx}
\usepackage{float}
\usepackage[ruled, vlined]{algorithm2e}
\usepackage{subfig}
\usepackage{color}
\usepackage{fullpage}
\usepackage{verbatim}
\usepackage{mkolar_definitions}
\usepackage{commath}
\usepackage{hyperref}
\usepackage{xspace}

\DeclareMathOperator{\sign}{sign}
\DeclareMathOperator{\logistic}{logistic}
\newcommand{\normal}{\textsc{Normal}\xspace}
\newcommand{\feedback}{\textsc{Feedback}\xspace}

\renewcommand{\paragraph}[1]{\vspace{1mm} \noindent{\bf #1}}

\def\shownotes{1}  \ifnum\shownotes=1
\newcommand{\authnote}[2]{$\ll$\textsf{\footnotesize #1 notes: #2}$\gg$}
\else
 \newcommand{\authnote}[2]{}
\fi

\title{SIM-ECG: A Signal Importance Mask-driven ECG Classification System
}

\author{Dharma KC\\
University of Arizona\\
\texttt{kcdharma@email.arizona.edu}
\And
Chicheng Zhang \\
University of Arizona\\
\texttt{chichengz@cs.arizona.edu}
\And
Chris Gniady \\
University of Arizona \\
\texttt{gniady@cs.arizona.edu}
\And
Parth Sandeep Agarwal \\
University of Arizona \\
\texttt{parthagarwal@email.arizona.edu}
\And
Sushil Sharma \\
\texttt{sushilamrahs@gmail.com}
}

\begin{document}
\maketitle
\begin{abstract}
Heart disease is the number one killer, and ECGs can assist in early diagnosis and prevention of deadly outcomes. Accurate ECG interpretation is critical in detecting heart diseases; however, they are often misinterpreted due to a lack of training or insufficient time spent to detect minute anomalies. Subsequently, researchers turned to machine learning to assist in the analysis. However, existing systems are not as accurate as skilled ECG readers,  and black-box approaches to providing diagnosis result in a lack of trust by medical personnel in given diagnosis. To address these issues, we propose a signal importance mask feedback-based machine learning system that continuously accepts feedback, improves accuracy and explains the resulting diagnosis. This allows medical personnel to quickly glance at the output and either accept the results, validate the explanation and diagnosis, or quickly correct areas of misinterpretation, giving feedback to the system for improvement. We have tested our system on a publicly available dataset consisting of healthy and disease-indicating samples. We empirically show that our algorithm is better in terms of standard performance measures such as F-score and Macro AUC compared to normal training baseline (without feedback); we also show that our model generates better interpretability maps.  
\end{abstract}

\section{Introduction}
Heart disease is the leading cause of death in the US and worldwide for both men and women~\cite{lancellotti2021cancer}.
In the US, heart diseases have been the leading cause of death since 2015, and the number of deaths from heart disease increased by 4.8\% from 2019 to 2020~\cite{ahmad2021leading}. The Center for Disease Control and Prevention notes that heart disease accounts for one in every four deaths in the United States each year. These deaths are attributed to many factors ranging from undetected heart diseases causing sudden death, late detection that may damage heart muscles and require repair, or even improper monitoring after successful heart surgery. Every year, more than 5 million Americans are affected by Heart Failures. Electrocardiogram (ECG), which records the heart's electrical activity, has long been the preferred and trusted technique for doctors to detect and diagnose these heart conditions. Therefore, accurate ECG analysis is critical in early disease detection and saving patients' lives. However, this is not the case quite frequently. A recent study found that 30\% of myocardial infarction events were misclassified as low risk, with ECG misinterpretation responsible for half of the misclassifications~\cite{FARAMAND2019161}. Misdiagnosis is also a top concern expressed by cardiac patients~\cite{cariacconcern}. It is therefore essential to develop mechanisms to improve analysis accuracy. 

The automation of ECG reading has been a long-standing need. Trying to analyze the signals manually is mundane, requires extreme concentration, causes mental fatigue, and is not well reimbursed adequately, to the amount of effort, by the insurance companies~\cite{drew2014finding}. Consequently, the number of people who can read ECG signals is shrinking, and experienced doctors don't have enough time to scrutinize patients' ECGs. However, commercial ECG machines only aid with basic preprocessing and analysis: they give information such as PR intervals, QRS duration, irregular rhythms, and a potential indication of abnormal ECG signals -- these analyses are not very sophisticated and offer limited assistance in detection of more complex cardiac conditions~\cite{hoffmann2020survey,velic2013computer}. Furthermore, many arising personal ECG monitoring devices also provide basic analysis either locally or send data to more sophisticated processing and eventually, human reading~\cite{steijlen2018novel}.

The machine learning community observed this need, and numerous machine learning algorithms were proposed for disease detection. Convolutional Neural Network assisted in arrhythmia and myocardial detection~\cite{kachuee2018ecg}. Other algorithms such as decision trees, $k$-nearest neighbor, logistic regression, support vector machines~\cite{haq2018hybrid}, and Inception neural networks~\cite{reasat2017detection} have also been evaluated. While existing machine learning algorithms have succeeded in classifying basic cardiac conditions, classification of more complex cardiac events is still challenging, and existing solutions are no match for highly skilled human readers. Furthermore, existing solutions provide diagnosis in a black-box manner, requiring the medicinal personnel to carefully analyze the ECG again to validate the algorithm's interpretations. We propose a system that can continuously collect ECG signal importance mask feedback from expert ECG readers and evolve its detection algorithms to surpass existing static solutions to address those challenges. Furthermore, the proposed system provides a visual interpretation of why the system decided and what portion of the signals were used in making the decision, allowing for quick validation by the reader. Our model and interpretability maps can save manual labor and provide readers with a precise region of the ECG signals where the disease manifestation is visible, which allows them to focus on validating the results instead of trying to find them in the first place.

\section{Related Work}
Deep learning has recently gained popularity in heart disease detection from 12-lead ECG signals. Models like residual convolutional neural networks, convolutional neural networks (CNNs), bidirectional long short term memory (LSTM), CNN and LSTM, CNNs, and densely-connected CNNs, have been used for small scale ($\leq12$-way) classification problem from 12-lead ECG signals ~\cite{ribeiro2020automatic, baloglu2019classification, mostayed2018classification, luo2019multi, hughes2018using, wang201912}. Lack of trust by medical personnel in black-box approaches resulted in recent efforts on interpretability analysis of ECG classification models~\cite{yao2020multi}, by using convolutional and recurrent neural networks for arrhythmia classification from 12-lead signals and providing explanation from attention mechanism. Similarly,~\cite{zhang2021interpretable} performs interpretability analysis using the Shapley additive explanations method. Recurrent neural networks (RNNs) for atrial fibrillation (AF) detection from a single-lead ECG signal combined with interpretability maps from attention mechanism have been explored~\cite{mousavi2020han}. All of the above methods are used for small scale classification ($\leq 12$-way) and do not learn from doctor's feedback. In contrast, our proposed solution simultaneously tackles large-scale (71-way) multi-label classification tasks, with a training procedure that considers doctors' signal importance masks (a set of time windows in ECG signal that are most indicative of the diseases exhibits) and provides model interpretability analysis.

Auxiliary information such as bounding boxes in computer vision~\cite{donahue2011annotator} and rationales in natural language processing~\cite{zaidan2008machine, sharma2018learning, zhang2016rationale} have been used to improve model accuracy and interpretability. Theoretical analysis on learning from auxiliary information beyond labels has also been explored~\cite{vapnik2009new, poulis2017learning, dasgupta2018learning}. Alignment of the attention maps generated by models to the attention maps labeled by experts to improve the accuracy and interpretability of CNNs has also been explored~\cite{mitsuhara2019embedding}. The addition of a regularization term in the training loss function that penalizes model gradients outside important regions in examples' input features has also been explored~\cite{ross2017right, dharmaimproving}, and we utilize those ideas in our solution. Use of part localization information to train CNNs to improve the accuracy has been proposed in~\cite{zhang2014part}. Recent use of explanation penalization terms to loss function that tries to match the model explanation with human explanation has improved the accuracy and interpretability of image and text classification systems~\cite{rieger2019interpretations}. Although using auxiliary information for improving the trustworthiness of image classification models and natural language models have been well studied in computer vision and natural language processing domains as discussed above, our work is the first that applies it for accurate and interpretable ECG signal-based heart disease prediction.

\section{System}
\label{sec:system}
The goal of our system is to detect cardiac diseases with high accuracy and provide its rationale, i.e., identifying important parts of the input ECG responsible for the predicted disease to allow quick validation by medical personnel and remove the black box approach of just providing the diagnosis. Existing solutions take labeled data to train models and use the trained models to classify new data in the field. This approach is desired if there is no expert to evaluate the data, and the system has to make decisions independently. Applying this classic approach, the doctors are presented with a result of a detected cardiac condition that they can accept or reject. This decision process is a black-box approach, and the doctor does not know how the machine arrived at the given classification. Our process starts similarly, but the learning system and doctor repeatedly communicate with each other with the following protocol: (1) Upon receiving a patient's ECG, the system detects the patient's heart condition while also pinpoints parts of the ECG that are responsible for its prediction -- for example, it highlights the difference in peaks in the ECG. This makes the decision process more trustworthy and acceptable to the medical community; (2) The doctors review the system's prediction and explanation and make their treatment decisions. The doctor can also provide feedback to the system by either confirming the prediction and explanation or correcting the prediction and by providing signal importance masks, i.e., parts of the ECG they believe are most relevant to the patient's condition.

\begin{figure}[H]
    \centering
    \includegraphics[width=\textwidth]{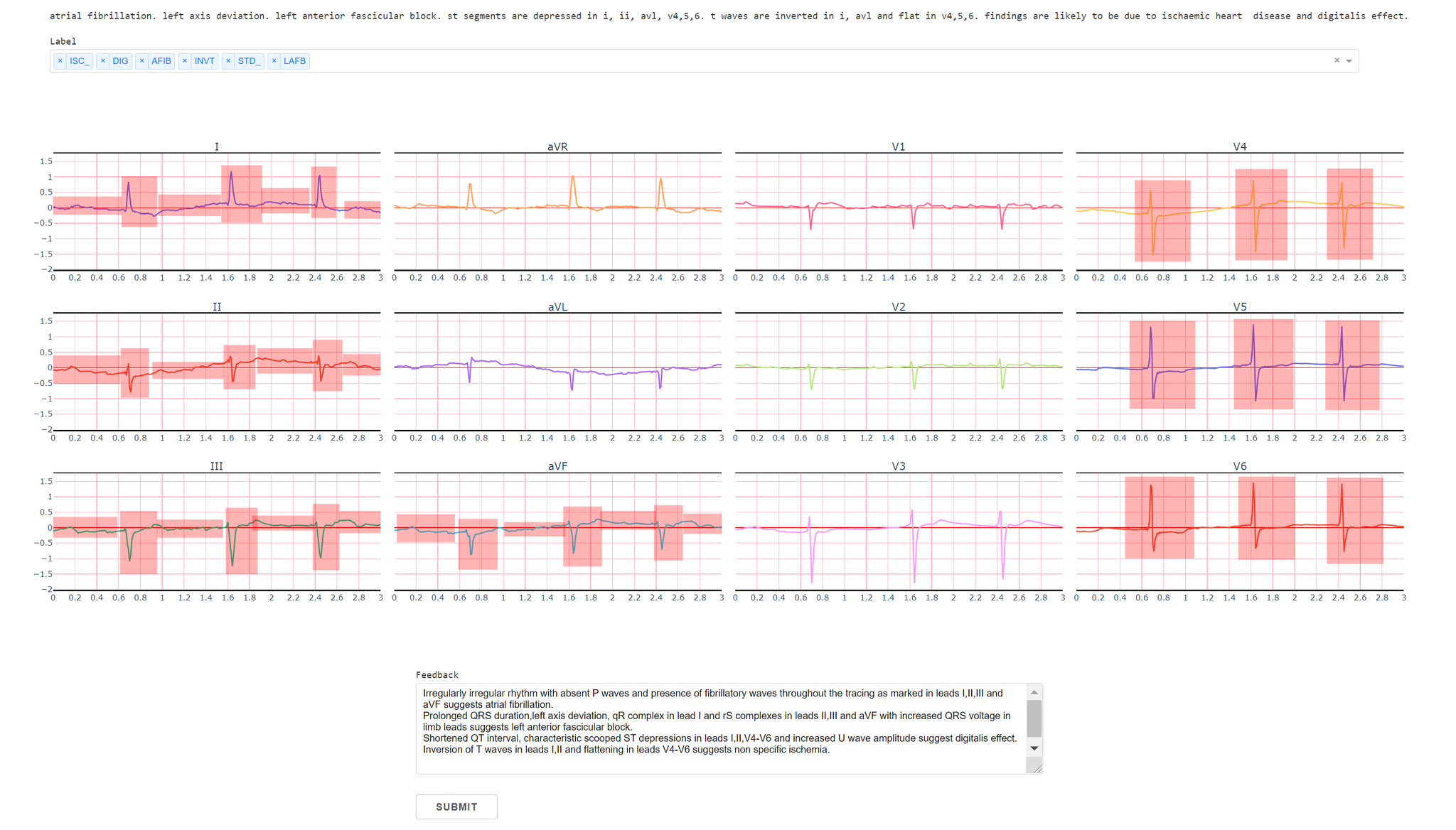}
    \caption{System interface for taking feedback and one example feedback}
    \label{fig:bounding_box}
\end{figure}

We created a web application where doctors can highlight important regions in 12-lead ECG signals to provide the signal importance masks. Figure~\ref{fig:bounding_box} shows our interface where doctors can highlight certain regions as well as provide feedback in the form of natural language explanation.
The web application was built using Flask with a PostgreSQL database. The doctors highlight all important regions for the given labels, and the feedback is stored in the database, which is subsequently used by our model training algorithm. Although we collect doctors' text explanations feedback in our system, we do not utilize them in our model training; we leave improving ECG classification models using text feedback as interesting future work.

\section{Model Training Algorithm}
\label{sec:algorithm}
Our model training algorithm is inspired by recent works on learning by incorporating interpretable auxiliary information~\cite{ross2017right}.
Recall that we would like to train classification models for 71-way heart disease classification from 12-lead ECG signals with signal importance masks being part of training data. Specifically, the training data are represented as a set of tuples $\cbr{ (x_i, y_i) }_{i=1}^n$, where for each example $i$, $x_i \in \RR^d$ is its 12-lead ECG signal representation; in our ECG dataset, $d=12\times300$. $y_i \in \cbr{-1,1}^{K}$ is the class labeling, where $K = 71$, for each coordinate $j \in \cbr{1,\ldots, K}$, $y_j = -1$ and $+1$ indicate that label $j$ is present and absent, respectively. 
In our training data, the doctor has provided signal importance masks $M_i \subset \cbr{1, \ldots, d}$ to some of the examples -- we denote by $E$ the index set of such examples.

Our model is represented by a function $f(x; \theta)$, where $x$ represents the ECG signal, and $\theta$ represents the model parameter. Given $x, \theta$, the model output $f(x; \theta)$ lies in $\RR^{K}$, and we use $\sign( f(x; \theta) ):= \del{ \sign(f(x; \theta)_1), \ldots, \sign(f(x; \theta)_K) } \in \cbr{-1,1}^K$ to denote its multi-label classification result, where $\sign(z) = 1$ if $z > 0$ and $\sign(z) = -1$ otherwise.
Our training process will hinge on finding $\theta$ that has a small average multi-label logistic loss on the training examples; here, the multi-label logistic
loss of model $f(\cdot; \theta)$ on a multi-label example $(x,y)$ is defined as 
$\ell_{\logistic}(\theta, (x,y)) = \sum_{j=1}^K \ln\del{1 + \exp(- y_j f_j(x; \theta))}$.

Our training objective function follows~\cite{ross2017right}
that proposes a regularized loss objective that takes advantage of signal importance masks, defined as:
\[
\min_\theta \del{ \sum_{i \in E} \ell_{\lambda}(\theta, (x_i, y_i, M_i)) +  \sum_{i \notin E} \ell_{\logistic}(\theta, (x_i, y_i)) },
\]
where 
\begin{equation} 
\ell_{\lambda}(\theta, (x, y, M) ) := 
\ell_{\logistic}(\theta, (x,y))
+  
\lambda \sum_{j \in [d] \setminus M} \rbr{ \pd{\ell_{\logistic}(\theta, (x,y))}{{x^j}} }^2,
\label{eqn:objective}
\end{equation}
and $\lambda > 0$ is a tuning parameter.
Each example $(x,y)$ can be viewed as introducing two parts of losses that contribute to the training objective: the first part is the standard multi-label logistic loss $\ell_{\logistic}(\theta, (x,y))$; the second part $\sum_{j \in [d] \setminus M} \rbr{ \pd{\ell_{\logistic}(\theta, (x,y))}{{x^j}} }^2$, only applies to examples in $E$ whose signal importance mask $M$ is available. This term regularizes the sensitivity of the model with respect to the input using the signal importance masks: it penalizes the model from being too sensitive to parts of the ECGs outside their corresponding signal importance masks.

\section{Experiments}
\label{sec:experiments}

We evaluate our algorithm using the large-scale 71-way multi-label heart disease classification dataset consisting of 12-lead ECG signals provided by~\cite{wagner2020ptb}. We follow the previously developed pipeline~\cite{strodthoff2020deep} as a baseline system with folds 1-8 as the training set, fold 9 as the validation set, and fold 10 as the test set. The dataset consists of 10-second recordings of the ECG signals with 17441 training examples, 2193 validation examples, and 2203 test examples. Similarly, we follow the previous evaluation metrics to evaluate and compare the models, based on Fmax and macro-AUC scores~\cite{strodthoff2020deep}. Macro AUC score is used to accommodate for class imbalance, as the number of normal ECG signals is quite high~\cite{wagner2020ptb}. We use the \texttt{inception1d} architecture to perform our experiments, which was shown to perform the best among all architectures~\cite{strodthoff2020deep}.

We use PyTorch~\cite{paszke2019pytorch} for all our experiments. Prior approach~\cite{strodthoff2020deep} uses multiple sliding windows of 2.5 seconds and aggregates the predictions from these windows to make a final prediction which acts as a data augmentation during the training phase. We use different training and inference to avoid the long time it takes for doctors to provide feedback for the full 10-second signal. We only use the first 3 seconds of the signal in training set for the feedback. To make a fair comparison between our model training and the baseline system, we use the first 3 seconds for training, but we still use full 10-seconds signals for validation and test.

We collected signal importance mask feedback from doctors for 1359 training examples out of all 17441 training examples. We use Adam optimizer with an initial learning rate of $0.002$ and batch size of 64 for all of our experiments. We repeated our experiment five times. 

\begin{figure}[H]
    \centering
    \includegraphics[width=\textwidth]{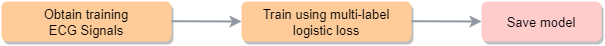}
    \caption{Normal training pipeline (\normal)}
    \label{fig:normal_training}
\end{figure}

\begin{figure}[H]
    \centering
    \includegraphics[width=\textwidth]{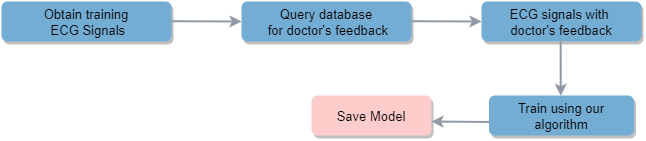}
    \caption{Our training pipeline  (\feedback)}
    \label{fig:our_training}
\end{figure}

Figure~\ref{fig:normal_training} shows the normal training pipeline (\normal for short), while Figure~\ref{fig:our_training} shows our training pipeline (\feedback for short).


\subsection{Results}
\begin{figure}[h]
    \subfloat[]{\includegraphics[width=3in, height=3in]{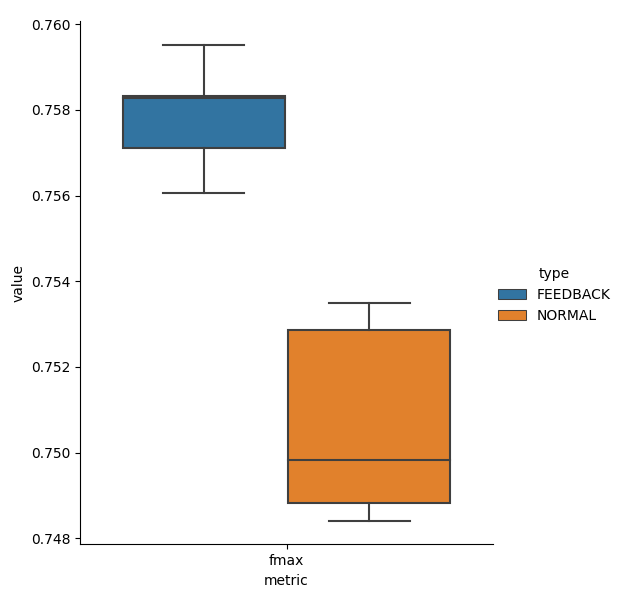}}
    \hspace{0.3in}
    \subfloat[]{\includegraphics[width=3in, height=3in]{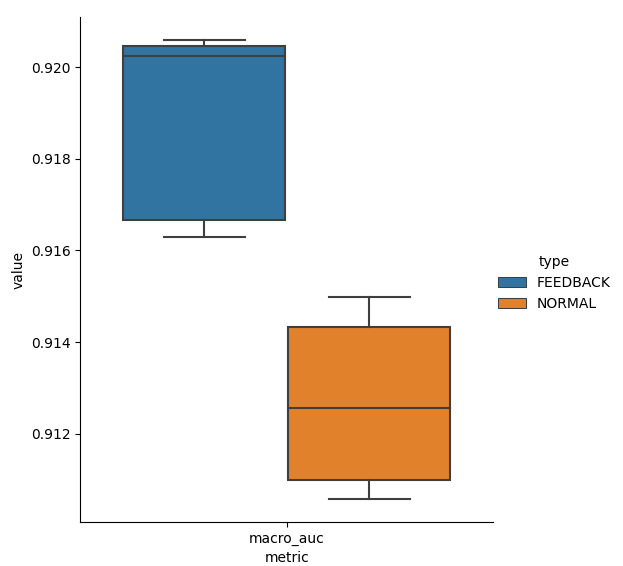}}
    \caption{Fmax boxplot and Macro AUC boxplot}
    \label{fig:fmax}
\end{figure}

Figure~\ref{fig:fmax} gives boxplots for Fmax and Macro AUC scores on the test dataset for \normal and \feedback. Thanks to incorporating doctors' feedback in training, \feedback has higher scores than \normal.

\begin{figure}[h]
    \centering
    \includegraphics[width=\textwidth]{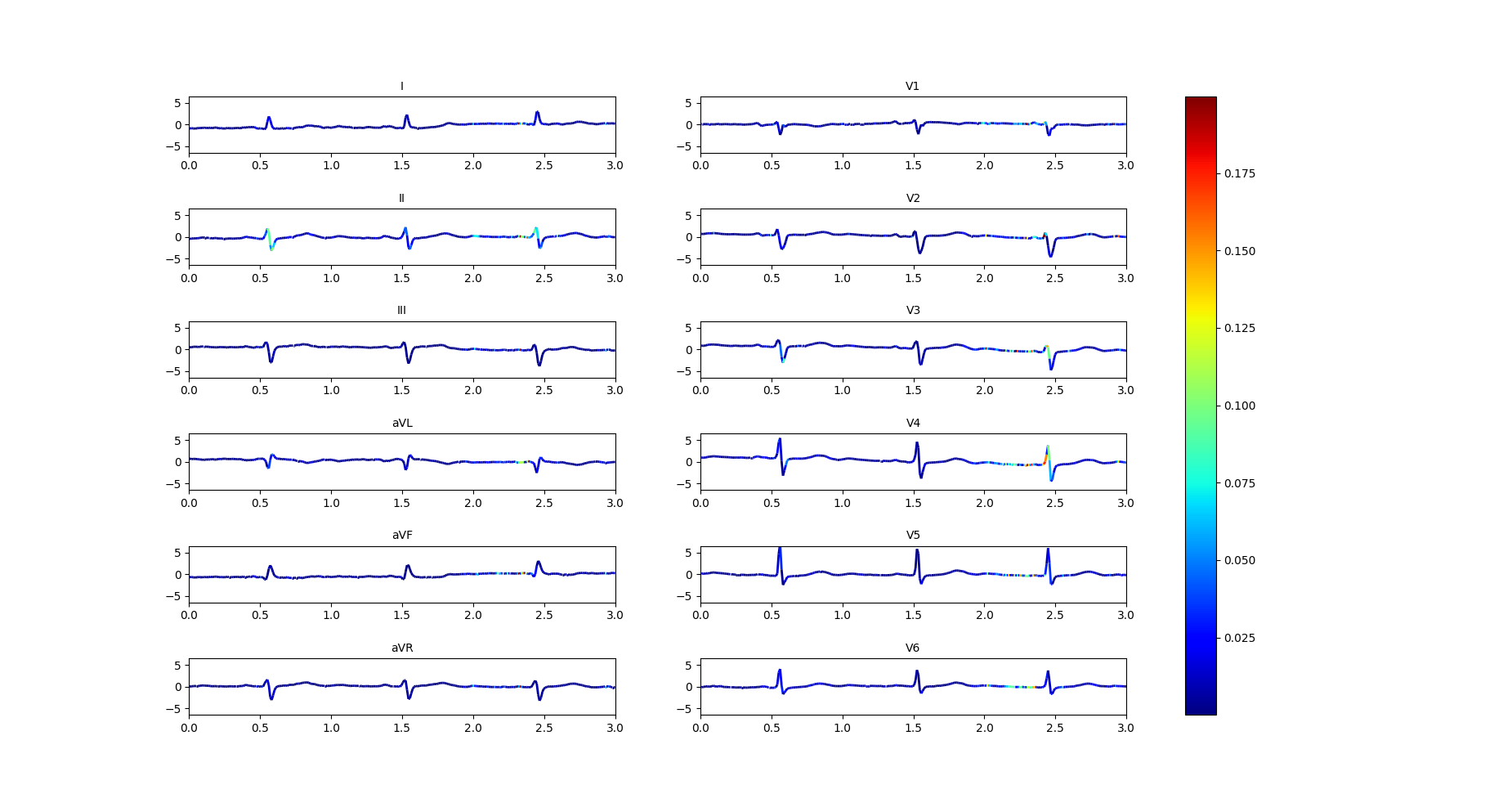}
    \caption{Interpretability map for a normal model}
    \label{fig:normal_grad}
\end{figure}

\begin{figure}[h]
    \centering
    \includegraphics[width=\textwidth]{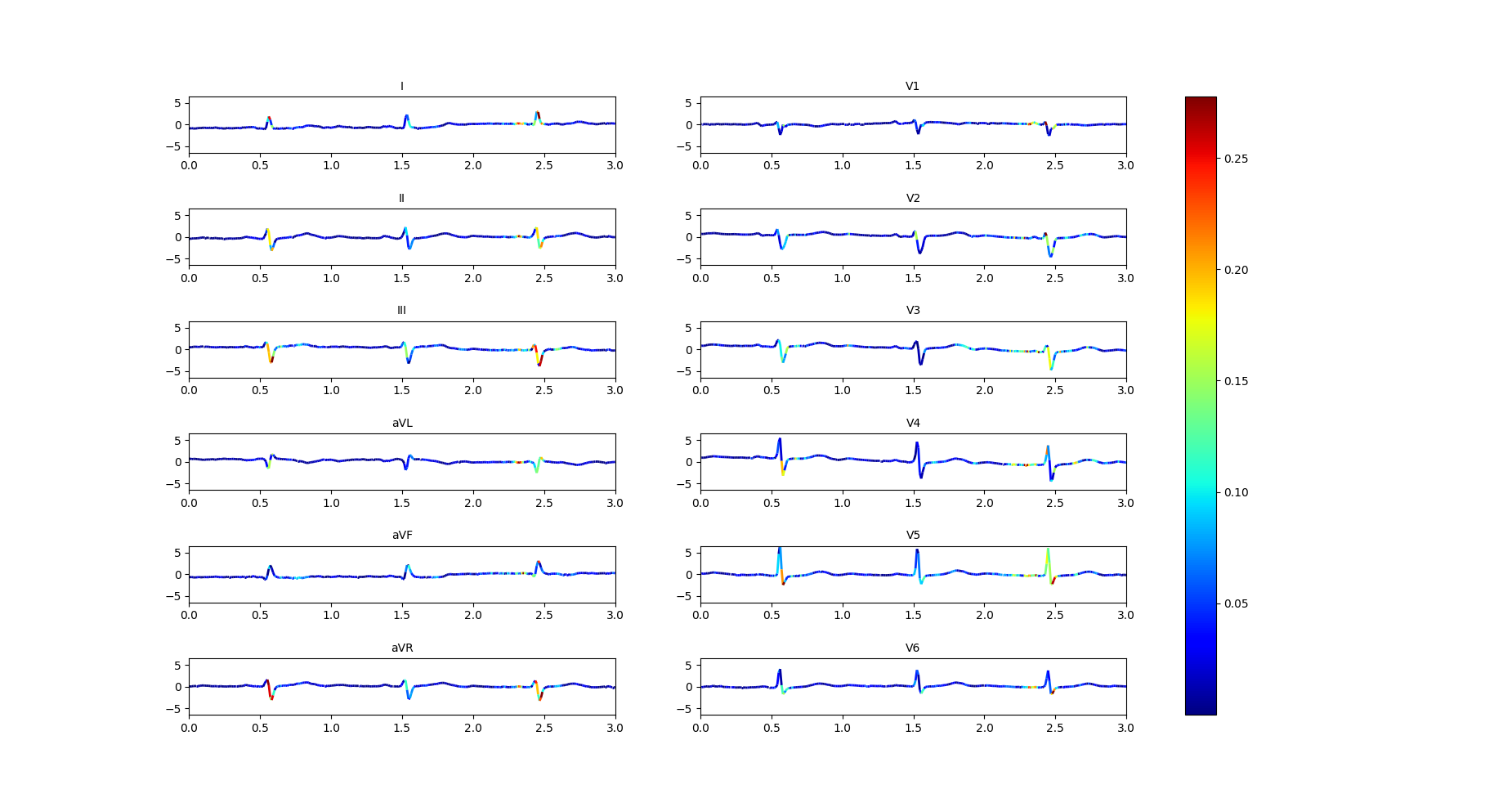}
    \caption{Interpretability map for a model trained with feedback}
    \label{fig:feedback_grad}
\end{figure}

Improvement in interpretability can accelerate ECG reading and evaluation by doctors and are at the core of our design. To generate the interpretability diagrams, we take the gradients of the output with respect to all 12 leads of the input signals and highlight the important regions~\cite{simonyan2013deep}, i.e., regions where the gradients are large.  The colormap is given in the right part of Figures~\ref{fig:normal_grad} and~\ref{fig:feedback_grad} where blue color (resp. red color) indicates lower (resp. higher) value of gradients, indicating regions of lower (resp. higher) importance; regions with colors that interpolate between blue and red indicates intermediate levels of importance. 

We have presented our interpretability maps to our medical team in a blind study. The interpretability maps generated by our models with feedback were always selected to be superior to those generated without the feedback. Figure~\ref{fig:normal_grad} and Figure~\ref{fig:feedback_grad} demonstrate this on an example with labels  `LAFB': Left Anterior Fascicular Block and `SR': Sinus Rhythm. Both models make correct predictions, but the model trained by \feedback highlights the correct regions while the model trained by \normal does not. Specifically, the doctor commented that: ``Left axis deviation, prolonged QRS duration, QR complexes in lead I and RS complexes in leads II, III, AVF and increased QRS voltages in these leads are more accurately highlighted in the model trained by \feedback to show left anterior fascicular block(LAFB). P waves are well marked to show a sinus rhythm (SR) as compared to the model trained by \normal, which doesn't highlight these important regions.''

\section{Conclusion}
We have built a system that can collect signal importance masks for 12-lead ECG signals from doctors. We have proposed and evaluated an ECG classification algorithm that utilizes signal importance mask feedback from doctors for accurate and interpretable predictions of 71 heart diseases. Our experiments demonstrate that collecting signal importance mask feedback for 1359 out of 17441 (7.7\%)  training examples already shows good improvement over the normal training baseline. We conjecture that training with larger feedback set can lead to even better models. In addition, we have collected natural language explanation feedback from doctors, which can be utilized in future works to train better models. We leave these as interesting avenues for future investigation.

\bibliographystyle{plain}
\bibliography{refs}

\end{document}